\documentclass{article}
\usepackage{spconf,amsmath,graphicx}

\usepackage{multirow}
\usepackage{tabularray}
\usepackage{tikz}
\usepackage{pgfplots}
\pgfplotsset{compat=newest}

\usepackage{subcaption}

\usepackage{multicol}
\usepackage{graphicx}

\title{Adapted Multimodal BERT with Layer-wise Fusion for Sentiment Analysis}
\name{Odysseas S. Chlapanis$^1$\quad Georgios Paraskevopoulos$^{1,2}$\quad Alexandros Potamianos$^{1}$}
\address{
        $^1$ National Technical University of Athens, Athens, Greece\\
        $^2$ Institute for Language and Speech Processing, Athena Research Center, Athens, Greece\\
}

\begin{document}
%
\maketitle
\begin{abstract}
Multimodal learning pipelines have benefited from the success of pretrained language models. However, this comes at the cost of increased model parameters. In this work, we propose Adapted Multimodal BERT (AMB), a BERT-based architecture for multimodal tasks that uses a combination of adapter modules and intermediate fusion layers. The adapter adjusts the pretrained language model for the task at hand, while the fusion layers perform task-specific, layer-wise fusion of audio-visual information with textual BERT representations. During the adaptation process the pre-trained language model parameters remain frozen, allowing for fast, parameter-efficient training. In our ablations we see that this approach leads to efficient models, that can outperform their fine-tuned counterparts and are robust to input noise. Our experiments on sentiment analysis with CMU-MOSEI show that AMB outperforms the current state-of-the-art across metrics, with $3.4\%$ relative reduction in the resulting error and $2.1\%$ relative improvement in $7-$class classification accuracy.

\end{abstract}

\section{Introduction}
\label{sec:intro}

Over the past few years, we have witnessed impressive breakthroughs in the field of multimodal applications, due to the abundance of multimedia data and progress in core machine learning algorithms. 
This has set the scene for multimodal machine learning as one of the frontiers of applied AI research. For wide-spread adoption in the real-world, models that strike the correct balance between performance and parameter efficiency should be developed.

GPT \cite{radford2018language} and BERT \cite{devlin-etal-2019-bert} were the first to establish the effectiveness of pre-training large scale language models on general tasks and then refining them for a specific task. Inspired by this approach, VilBERT \cite{lu_vilbert_2019} leveraged parallel multimodal data for pre-training a visual-language model. Other researchers \cite{tsimpoukelli2021multimodal, eichenberg_magma_2021, 10.1145/3394171.3413678, rahman_integrating_2020} have adopted a more flexible method: adapting a model pre-trained only on language for multimodal tasks.  

\begin{figure}[t]
\centering
\definecolor{frozen_color}{RGB}{33, 169, 237}
\definecolor{red_color}{RGB}{247, 56, 114}
\definecolor{orange_color}{RGB}{247, 136, 56}
\definecolor{purple_color}{RGB}{39, 186, 75}
\begin{tikzpicture}[scale=0.75]
\begin{axis}[
	xlabel={Trainable Parameters},
	ylabel={Accuracy 7class},
	mark size={5},
 	extra y ticks={51.8, 52.2, 53.3},
 	ymajorgrids,
 	xmajorgrids,
]

\addplot[color=purple_color, mark=square*, mark size=3pt,
        nodes near coords= \;MulT (G),
        every node near coord/.style={anchor=180}
] coordinates{
    (1.892441,51.8)
};

\addplot[color=purple_color, mark=square*, 
        mark size=3pt,
        nodes near coords= \;MMLatch (G),
        every node near coord/.style={yshift=+4pt,anchor=230} 
        ] coordinates{
    ( 2.600000,52.1)
};

\addplot[color=orange_color, mark=pentagon*,
        mark size=3pt,
        nodes near coords= \;TFN (B),
        every node near coord/.style={yshift=-2pt,anchor=120} 
        ] coordinates{
    (0.6,51.8)
};
\addplot[color=orange_color, mark=pentagon*,
        mark size=3pt,
        nodes near coords= \;LMF (B),
        every node near coord/.style={anchor=west} 
        ] coordinates{
    (1.0,50.2)
};
\addplot[color=orange_color, mark=pentagon*,
        mark size=3pt,
        nodes near coords= \;MFM (B),
        every node near coord/.style={yshift=-2pt,anchor=170} 
        ] coordinates{
    (1.7,51.3)
};
\addplot[color=red_color, mark=triangle,
            very thick, 
            mark size=4pt,
            nodes near coords= MISA (A),
            every node near coord/.style={xshift=3pt, anchor=180, font=\large}
        ] coordinates{
    (8.543377,52.15)
};
\addplot[color=red_color, mark=triangle*,
        mark size=4pt,
        nodes near coords= MISA (FT),
        every node near coord/.style={xshift=-8pt, yshift=+5pt,anchor=320, font=\large}
        ] coordinates{
    (47.124625,52.2)
};
\addplot[color=frozen_color, mark=*,
        mark size=4pt,
        nodes near coords= \textbf{AMB (A)},
        every node near coord/.style={xshift=2pt, yshift = -5pt, anchor=175, font=\Large}
        ] coordinates{
    (8.593231, 53.29)
};
\addplot[color=frozen_color, mark=o, 
        very thick, 
        mark size=4pt,
        nodes near coords= AMB (FT),
        every node near coord/.style={xshift=-6pt, yshift=-5pt, anchor=40, font=\large}
        ] coordinates{
    (47.174479, 51.98)
};
\end{axis}
\end{tikzpicture}
\caption{7-class accuracy with respect to number of trainable parameters for the best performing models in the literature. G stands for GloVe embeddings, A for adapters, B for frozen and FT for fine-tuned BERT embeddings. The proposed AMB with adapters achieves a good balance between trainable parameters and performance.}
\vspace{-1em}
\label{fig:parameters-lit}
\end{figure}
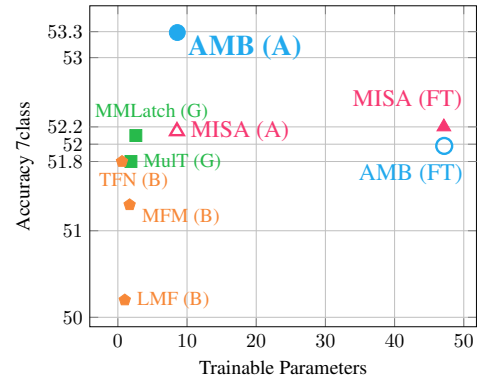

The standard method of transferring a pre-trained model to a downstream task is called fine-tuning, which involves updating the pre-trained weights with backpropagation. However this method incurs intensive data and computational costs, while some information is lost due to using only task-specific data for updating model parameters. This phenomenon is known as catastrophic forgetting \cite{bower_catastrophic_1989}.
To solve these issues, GPT-3 \cite{brown_language_2020} proposed ``prompt tuning", an intuitive method to transfer a powerful pretrained model only with text interactions, called ``prompts", without any gradient updates.
This idea was later extended, with many variations \cite{lester-etal-2021-power, li-liang-2021-prefix}, to make these prompts trainable, now called ``soft prompts". Houlsby et al. \cite{houlsby2019parameter} proposed adapters, a down-projected feedforward network that updates the representations of each BERT layer. 
Frozen \cite{tsimpoukelli2021multimodal} applied these ideas in multimodal learning, by translating an image to a visual soft prompt that is prepended to the input of a standard language model, which keeps its original pre-trained weights  unchanged (frozen).
MAGMA \cite{eichenberg_magma_2021} extended this by showing that the addition of adapter layers \cite{houlsby2019parameter} in between the frozen language layers outperforms Frozen.
Flamingo \cite{alayrac_flamingo_2022} scaled up and optimised this concept by introducing a flexible visual encoder which can turn arbitrary sequences of images or even video frames to a fixed number of visual tokens.

Early applications of deep learning for multimodal sentiment analysis focused on the use of Recurrent Neural Networks (RNNs) \cite{metallinou_context-sensitive_2015, wollmer_lstm-modeling_2013, shenoy-sardana-2020-multilogue} and Convolutional Neural Networks (CNNs) \cite{poria_convolutional_2016} aiming to model contextual information.
The next innovation was the introduction of the attention model to create sophisticated fusion approaches \cite{gu_multimodal_2018, wang_words_2018}.
This naturally led to the incorporation of the transformer \cite{vaswani_attention_2017} as the central model for this task \cite{DBLP:journals/corr/abs-1806-06176, delbrouck-etal-2020-transformer}.
Lately, large-scale pretrained language transformers, such as BERT \cite{devlin-etal-2019-bert}, have become the norm because of consistent performance gains.
ICCN \cite{Sun2020LearningRB} introduced Deep Canonical Correlation Analysis for jointly learning representations.
Wang et al. \cite{wang_words_2018} and later MAG-BERT \cite{rahman_integrating_2020} proposed shifting methods.
MISA \cite{10.1145/3394171.3413678} produced modality invariant and modality specific representations in an effort to disentangle data relationships.
More recently, many researchers turned their efforts towards intricate multimodal pre-training
strategies, such as \cite{yu2021le, kim_cmsbert-clr_2022}. Such methods are model-agnostic and should be studied separately for a fair comparison.

We present a simple neural architecture that adapts BERT representations for multimodal fusion which we call Adapted Multimodal BERT (AMB). Our approach extends concepts introduced by visual-language models \cite{tsimpoukelli2021multimodal,eichenberg_magma_2021, alayrac_flamingo_2022} to include audio.
The contributions of our work:
\begin{itemize}
    \item AMB is evaluated on multimodal sentiment analysis with the CMU-MOSEI database to achieve new state-of-the-art results, regardless of being lightweight and data-efficient due to a low trainable parameter budget.
    \item BERT is tuned in an effective way to adapt without losing prior knowledge, while at the same time squeezing as much useful information as possible from audio-visual modalities.
    \item We study our model's robustness to noise and compare its performance with a fine-tuned version and the current state-of-the-art MISA.
\end{itemize}

\section{Proposed Method}
\label{sec:method}

\begin{figure}[t]
\centering
\includegraphics[scale=0.8]{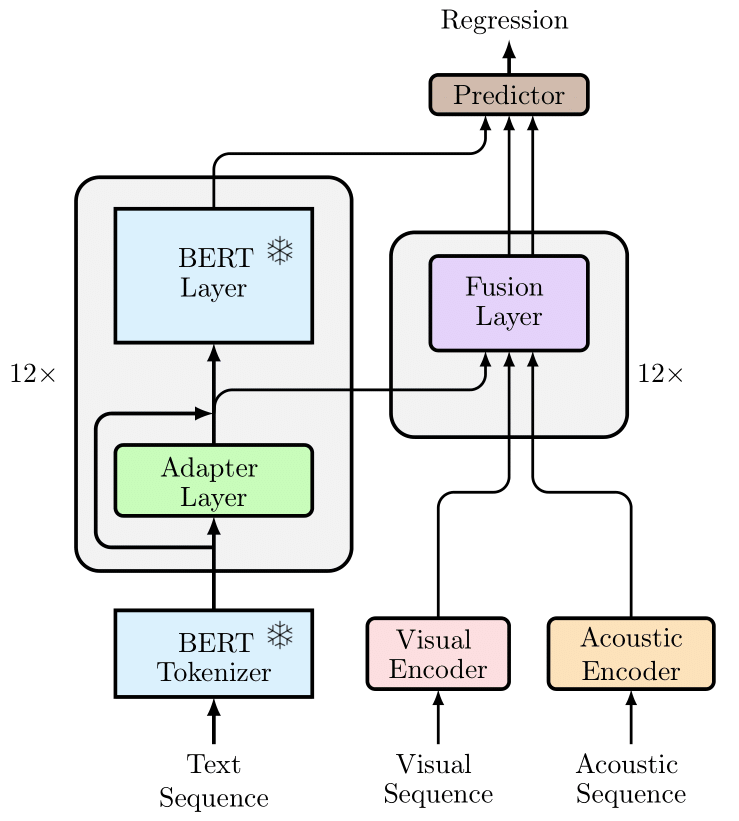}
\caption{Architecture of Adapted Multimodal BERT (AMB)}
\label{fig1:architecture}
\end{figure}

Fig.~\ref{fig1:architecture} illustrates an overview of the system architecture. 
First of all, the input sequences are fed into their respective encoders to prepare for the next stage.
The core component is a frozen pre-trained BERT model, which is tuned by adapter layers, without access to any other modalities.
These BERT representations are combined with audio-visual information in a feedforward network (FFN) in order to perform layer-wise multimodal fusion. 
This process is repeated for 12 layers and the last representations are provided to a FFN to predict the sentiment score.


\noindent\textbf{Frozen BERT layers:}
The frozen BERT model is at the core of the architecture to emphasize the importance of language. Both BERT tokenizer and the $12$ BERT layers are kept intact during training, limiting the effects of catastrophic forgetting that can incur during fine-tuning. 

\noindent\textbf{Adapter layers:}
We use the original bottleneck adapters, introduced by Houlsby et al. \cite{houlsby2019parameter}.
Each adapter layer is composed of a linear down-projection followed by a ReLU non-linearity and then a linear up-projection to restore the original input dimensions. Residual connections are used between the input and output of each adapter layer.
Instead of inserting an adapter layer both between the attention and the feedforward module, we follow \cite{pfeiffer-etal-2021-adapterfusion} and only insert them after the feedforward layernorm layers, thus cutting the number of additional parameters in half.
Our adapter layers are only responsible for adapting to the textual inputs.

\noindent\textbf{Visual and Audio Encoders:}
Visual and audio encoders consist of transformer encoder layers that act on each modality separately to extract information from an arbitrary sequence of features and compress it in a concatenated visual-acoustic token. This token is then prepared for the next stage of layer-wise multimodal fusion. Our encoders are closely related to the approach of \cite{tsimpoukelli2021multimodal, eichenberg_magma_2021, alayrac_flamingo_2022}, with the addition of audio.

\begin{table*}
  \centering
  \begin{tblr}{ |c|c|c|c|c|c|c|}
    \hline
    Models & MAE $(\downarrow)$ & Corr $(\uparrow)$ & Acc-7 $(\uparrow)$ & Acc-2 $(\uparrow)$ & F1 $(\uparrow)$ & Trainable Parameters\\\hline
    MMLatch (G) \cite{9746418} & $0.582$ & $0.704$ & $52.1$ & $82.8$ & $82.9$ &  $2.6$\\
    MulT (G) \cite{tsai-etal-2019-multimodal} & $0.580$ & $0.703$ & $51.8$ & $82.5$ & $82.3$ &  $1.8$\\
    \hline
    LMF (B) \cite{liu-etal-2018-efficient-low} & $0.623$ & $0.677$ & $50.2$ & $82.0$ & $82.1$  & $1.0$\\
    TFN (B) \cite{tensoremnlp17} & $0.593$ & $0.700$ &   $51.8$ & $82.5$ & $82.3$ & $0.6$\\
    MFM (B) \cite{DBLP:journals/corr/abs-1806-06176} & $0.568$ & $0.717$ & $51.3$ & $84.4$ & $84.3$ & $1.7$\\
    ICCN (B) \cite{Sun2020LearningRB}  & $0.565$ & $0.713$ & $51.6$ & $84.2$ & $84.2$  &  $-$\\
    \hline
    MAG-BERT$^*$ (FT) \cite{rahman_integrating_2020} & $0.614$ & $0.763$ & $50.9$ & $84.3$ & $84.2$ &  $110.8$\\
    MISA (FT) \cite{10.1145/3394171.3413678} & $0.555$ & $0.756$ & $52.2$ & $85.3$ &  $85.3$  &  $47.1$\\
    \hline 
    AMB (Ours) & $\mathbf{0.536}$ & $\mathbf{0.766}$ & $\mathbf{53.3}$ & $\mathbf{85.8}$ & $\mathbf{85.8}$ & $8.6$\\
 \hline
  \end{tblr}
   \caption{Results on CMU-MOSEI. Models indicated with (G) use glove embeddings. Models indicated with (B) use frozen BERT embeddings, and are taken from \cite{Sun2020LearningRB}. MISA and MAG-BERT use a fine-tuned (FT) BERT for feature extraction from language.  MAG-BERT$^*$ is reproduced for CMU-MOSEI by the authors of this paper. Trainable parameters are in millions.}
   \label{tab:sota}
\end{table*}

\noindent\textbf{Fusion layers:}
For multimodal fusion FeedForward Network Fusion (FFN-Fusion) is used in a layer-wise manner, between each BERT layer. The first BERT token (known as CLS token), which is commonly used to store a semantic summary of BERT's hidden states \cite{lu_vilbert_2019}, is projected to a lower dimension and then concatenated with the modality tokens produced by the visual and audio encoders. This tensor is then fed into FFN-Fusion to output the fused representations. Although \cite{rahman_integrating_2020} and \cite{alayrac_flamingo_2022} also perform layer-wise multimodal fusion, both use the result to shift BERT representations in order to generate output text. We adopt a simpler approach without shifting.


\noindent\textbf{Predictor:} The fused representation of the last BERT and fusion layers are concatenated and fed into a classification head, consisting of a single Feedforward layer. Minimum Absolute Error loss is used for end-to-end training of the network.

\section{Experimental Setup}

\noindent\textbf{Data: }
The proposed model is evaluated for sentiment analysis on CMU-MOSEI \cite{Zadeh2018MultimodalLA}.
It contains $23,454$ YouTube video clips of reviews on movies or other topics, where each sample is manually annotated with a sentiment score, ranging from $-3$ (strongly negative) to $+3$ (strongly positive).
Text transcriptions are segmented into words, while visual FACET and acoustic COVAREP features are collected and aligned on these words. Standard train, development and test splits are provided.
For evaluation, mean absolute error (MAE) and Pearson Correlation (Corr) between model and human predictions are used for regression, while seven-class accuracy (Acc-7), binary accuracy (Acc-2) and F1-score (F1) are used for classification.

\noindent\textbf{Implementation Details:}
\label{ssec:setup}
The \texttt{bert-base-uncased} version of BERT \cite{devlin-etal-2019-bert} is used for all experiments.
It contains $12$ transformer layers, where each token of the sentence has hidden size of $768$ dimensions.
The tokens are prepared for BERT with the standard tokenization procedure, while the two special tokens, [CLS] and [SEP], are added at the start and in the end of each sentence respectively.
The encoders used for visual and acoustic modalities are randomly initialized transformer encoder modules with $2$ layers and $1$ attention head. 
We find that prepending a learnable [CLS] token and collecting this as a semantic summary works best.
After a short hyper-parameter search in the range [128, 768] for the hidden size of the adapter layers, $384$ is chosen as the optimal value. Similarly, for fusion layers $220$ is chosen from [160, 820] as the hidden size.

For optimization, the Adam optimizer 
\cite{DBLP:journals/corr/KingmaB14} 
is used with learning rate $5*10^{-5}$. Early stopping is used with patience set to $10$ epochs and dropout is set to $0.2$. Training takes $20$ minutes on a single GTX 1080Ti NVIDIA GPU.

\section{Experiments and Results}

\label{sec:experiments}
\begin{figure*}[h!]
\centering
\begin{multicols}{3}
        \begin{subfigure}{0.33\textwidth}
            \includegraphics[scale=0.65]{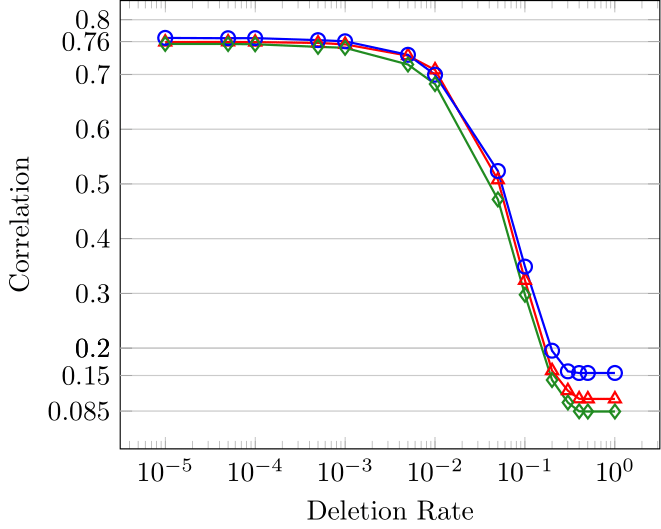}
            \label{fig:delete}
        \end{subfigure}
        \hfill
        \begin{subfigure}{0.33\textwidth}
            \includegraphics[scale=0.65]{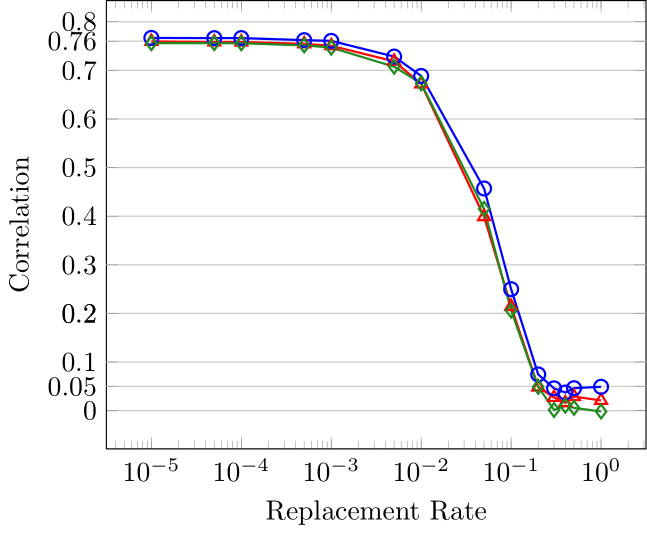}
            \label{fig:replace}
        \end{subfigure}
        \hfill
        \begin{subfigure}{0.33\textwidth}
        \includegraphics[scale=0.65]{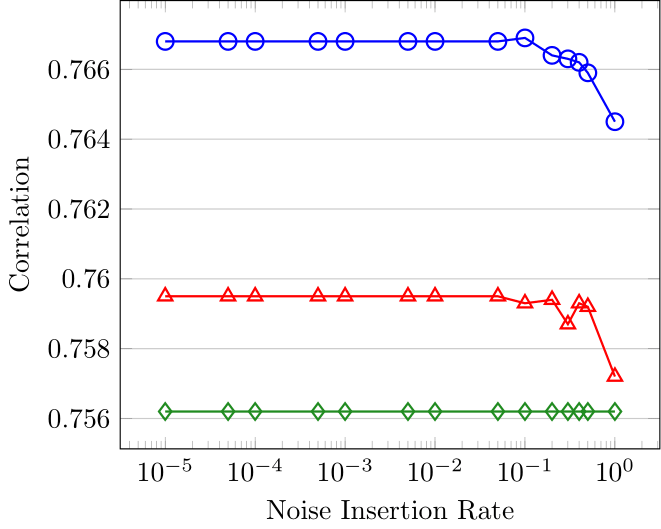}
        \label{fig:visual}
    \end{subfigure}
    \end{multicols}
\vspace{-1.5em}
\caption{Model robustness for varying levels of noise, i.e. random deletion of input tokens (left), random replacement of input tokens (middle), noise insertion to visual inputs (right). Blue $^\circ$: AMB, Red $^\triangle$: MISA, Green $^\diamond$: AMB-FT}
\label{fig:robustness}
\end{figure*}

\begin{table}[t!]
  \centering
  \small
\begin{tblr}{|c|c|c|c|}
    \hline
     Models & Corr $(\uparrow)$ &  Acc-7 $(\uparrow)$ &  Train. Params \\\hline
       AMB no-text & $0.240$ & $41.64$   & $8.6$\\ 
    AMB text-only &  $0.760$ & $52.81$ & $8.6$\\ 

 \hline  
    MISA-Adapters &  $0.758$ & $52.15$ & $8.5$\\ 
    MISA & $0.756$ & $52.20$ & $47.1$\\ 
    \hline

    AMB-FT & $0.756$ & $51.98$ & $47.2$\\ 
    \textbf{AMB}  & $\mathbf{0.766}$ & $\mathbf{53.29}$ & $8.6$ \\ 
 \hline
  \end{tblr}
  \caption{Multimodal adapters vs fine-tuning. We include experiments, where the text, or the audio-visual modalities are missing. Trainable parameters are in millions.}\label{tab:adapters-vs-ft}
  \vspace{-1em}
\end{table}

\noindent\textbf{Comparison to state-of-the-art:}
\label{ssec:results}
The results for multimodal sentiment analysis on CMU-MOSEI are presented in Table~\ref{tab:sota}. 
For fair comparison we only compare with methods in the literature that train in one stage, without leveraging their own, separate, pre-training stage on multimodal data. We observe that the proposed model outperforms all other methods by a significant margin across all metrics.
As shown in Fig.~\ref{fig:parameters-lit}, models that utilize Glove embeddings (G) \cite{pennington2014glove}, or frozen BERT embeddings (B), have fewer trainable parameters, sacrificing overall performance. Models that rely on fine-tuning of BERT have a significantly larger amount of trainable parameters. AMB with adapters surpasses fine-tuning based approaches on a small parameter budget.

\noindent\textbf{Ablation studies:}
\label{ssec:adapters vs ft}
Table~\ref{tab:adapters-vs-ft} shows an ablation study on the effect of the exclusion of modalities and the effect of using adapters versus finetuning for the adaptation of the language model.
Firstly, the exclusion of the textual modality significantly degrades performance for the ``AMB no-text'' model, which demonstrates that text is the dominant modality for this task. With the exclusion of audio-visual information in ``AMB text-only'' preformance still declines, though to a lesser degree, indicating that the use of multimodal information is beneficial.

For the adapters versus fine-tuning experiments, an adapter based version of MISA (``MISA-Adapters'') and a fine-tuned version of AMB (``AMB-FT'') are implemented. We observe that fine-tuning is either unnecessary as in the case of MISA or even decreases model performance as in the case of AMB, revealing that some catastrophic forgetting occurs when performing fine-tuning on the text modality in this multimodal setting.

\noindent\textbf{Noise Robustness:}
\label{ssec:robustness}
Finally, we evaluate the robustness of our model with respect to noise insertion in the visual and text modalities.
When testing the robustness for the visual modality, we follow Hazarika et al. \cite{hazarika2022analyzing}, who propose the insertion of multiplicative Gaussian noise to a randomly selected set of input sequence elements for a given modality. For the text modality a different approach is employed that more closely simulates real-world errors, i.e. deleting and replacing input tokens. In the token replacement experiment a percentage of input tokens is selected randomly and replaced with random tokens from the vocabulary, while for the token deletion experiment they are instead replaced with the [UNK] token. We select the best checkpoint of each model and show the average correlation over three independent runs, following \cite{hazarika2022analyzing}.

Fig.~\ref{fig:robustness} displays the results of the robustness tests for varying levels of input noise. The deletion, replacement and noise insertion rate refer to the probability of corrupting each element in the input sequence.
When corrupting textual inputs by deleting or replacing tokens we observe that performance starts to degrade after corrupting each token with $5\%$ probability. Steeper performance degradation occurs in the case of replacement than in the case of deletion. This sensitivity to noise is expected, as text is the dominant modality. We observe similar robustness characteristics for AMB, MISA and AMB-FT, though adapter-based AMB appears to be somewhat more robust than its fine-tuned counterpart. 
In the extreme case from $50\%$ probability and beyond AMB's lowest point is significantly higher than the rest, verifying that it considers all modalities to make predictions.  
In the case of noise injection to the visual modality performance drops off for AMB and MISA at $10\%$ noise insertion rate. We observe that noise insertion in the visual modality affects both models less than noise insertion in text. Interestingly, the AMB-FT model is not affected by visual noise, revealing that this model relies completely on text, ignoring visual cues. 
These results highlight that, favoring adapter-based approaches over fine-tuning when using large pre-trained language models for multimodal tasks may lead to improved model robustness and better utilization of information from less dominant modalities (that contribute less to overall performance).



\section{Conclusions}
\label{sec:conclusions}
In this work, AMB is proposed, a simple yet innovative model that builds on a powerful pre-trained BERT transformer encoder and avoids the pitfalls of catastrophic forgetting and modality imbalance, i.e., useful knowledge from pre-training and non-dominant modalities is leveraged effectively.
Further, the use of adapters allows our model to lower the cost of trainable parameters and leads to improved robustness to various types of noise.

In the future, we plan to extend our experiments to more tasks, such as text generation from input prompts enriched with images. Moreover, exploring more sophisticated fusion methods compatible with our approach might be beneficial. The effects of shifting should also be considered.
We hope that this approach will be viewed as the blueprint for designing multimodal models based on pre-trained unimodal encoders in a flexible and effective manner. 


\small
\bibliographystyle{IEEEbib}
\bibliography{strings,refs}

\begin{thebibliography}{10}

\bibitem{radford2018language}
Radford et~al.,
\newblock ``Improving language understanding by generative pre-training,''
\newblock 2018.

\bibitem{devlin-etal-2019-bert}
J.~Devlin, M.~W. Chang, K.~Lee, and K.~Toutanova,
\newblock ``{BERT}: Pre-training of deep bidirectional transformers for
  language understanding,''
\newblock in {\em Proc. NAACL, Vol. 1}. 2019, pp. 4171--4186, ACL.

\bibitem{lu_vilbert_2019}
J.~Lu et~al.,
\newblock ``{ViLBERT}: {Pretraining} {Task}-{Agnostic} {Visiolinguistic}
  {Representations} for {Vision}-and-{Language} {Tasks},''
\newblock in {\em NeurIps}, 2019, vol.~32.

\bibitem{tsimpoukelli2021multimodal}
Maria Tsimpoukelli et~al.,
\newblock ``Multimodal few-shot learning with frozen language models,''
\newblock in {\em NeurIps}, A.~Beygelzimer et~al., Eds., 2021.

\bibitem{eichenberg_magma_2021}
C.~Eichenberg et~al.,
\newblock ``{MAGMA} -- {Multimodal} {Augmentation} of {Generative} {Models}
  through {Adapter}-based {Finetuning},''
\newblock {\em arXiv:2112.05253}, 2021.

\bibitem{10.1145/3394171.3413678}
D.~Hazarika, R.~Zimmermann, and S.~Poria,
\newblock ``Misa: Modality-invariant and -specific representations for
  multimodal sentiment analysis,''
\newblock in {\em Proc. 28th ACM}. 2020, MM '20, p. 1122–1131, ACM.

\bibitem{rahman_integrating_2020}
W.~Rahman et~al.,
\newblock ``Integrating {Multimodal} {Information} in {Large} {Pretrained}
  {Transformers},'' 2020,
\newblock arXiv:1908.05787.

\bibitem{bower_catastrophic_1989}
M.~McCloskey and N.~J. Cohen,
\newblock ``Catastrophic {Interference} in {Connectionist} {Networks}: {The}
  {Sequential} {Learning} {Problem},''
\newblock Academic Press, 1989,
\newblock ISSN: 0079-7421.

\bibitem{brown_language_2020}
Brown et~al.,
\newblock ``Language {Models} are {Few}-{Shot} {Learners},'' 2020,
\newblock arXiv:2005.14165 [cs].

\bibitem{lester-etal-2021-power}
B.~Lester, R.~Al-Rfou, and N.~Constant,
\newblock ``The power of scale for parameter-efficient prompt tuning,''
\newblock in {\em Proc. 2021 EMNLP}. 2021, pp. 3045--3059, ACL.

\bibitem{li-liang-2021-prefix}
X.~L. Li and P.~Liang,
\newblock ``Prefix-tuning: Optimizing continuous prompts for generation,''
\newblock in {\em Proc. 59th Annual Meeting of the ACL}. 2021, pp. 4582--4597,
  ACL.

\bibitem{houlsby2019parameter}
N.~Houlsby, A.~Giurgiu, et~al.,
\newblock ``Parameter-efficient transfer learning for {NLP},''
\newblock in {\em Proc. 36th ICML}, 2019.

\bibitem{alayrac_flamingo_2022}
J.~B. Alayrac et~al.,
\newblock ``Flamingo: a {Visual} {Language} {Model} for {Few}-{Shot}
  {Learning},''
\newblock Tech. {R}ep., 2022,
\newblock arXiv:2204.14198.

\bibitem{metallinou_context-sensitive_2015}
A.~Metallinou et~al.,
\newblock ``Context-sensitive learning for enhanced audiovisual emotion
  classification ({Extended} abstract),''
\newblock in {\em 2015 {ACII}}, 2015.

\bibitem{wollmer_lstm-modeling_2013}
M.~Wöllmer et~al.,
\newblock ``{LSTM}-{Modeling} of continuous emotions in an audiovisual affect
  recognition framework,''
\newblock {\em Image and Vision Computing}, vol. 31, pp. 153--163, 2013.

\bibitem{shenoy-sardana-2020-multilogue}
A.~Shenoy et~al.,
\newblock ``Multilogue-net: A context-aware {RNN} for multi-modal emotion
  detection and sentiment analysis in conversation,''
\newblock in {\em Challenge-HML}. 2020, pp. 19--28, ACL.

\bibitem{poria_convolutional_2016}
S.~Poria et~al.,
\newblock ``Convolutional {MKL} {Based} {Multimodal} {Emotion} {Recognition}
  and {Sentiment} {Analysis},''
\newblock in {\em 2016 {IEEE} 16th {ICDM}}, 2016, pp. 439--448.

\bibitem{gu_multimodal_2018}
Y.~Gu et~al.,
\newblock ``Multimodal {Affective} {Analysis} {Using} {Hierarchical}
  {Attention} {Strategy} with {Word}-{Level} {Alignment},''
\newblock in {\em Proc. 56th ACL}. 2018, pp. 2225--2235, ACL.

\bibitem{wang_words_2018}
Y.~Wang et~al.,
\newblock ``Words can shift: Dynamically adjusting word representations using
  nonverbal behaviors.,''
\newblock 2019, pp. 7216--7223, AAAI.

\bibitem{vaswani_attention_2017}
A.~Vaswani, N.~Shazeer, et~al.,
\newblock ``Attention is {All} you {Need},''
\newblock in {\em NeurIps}, I.~Guyon et~al., Eds. 2017, vol.~30, Curran
  Associates, Inc.

\bibitem{DBLP:journals/corr/abs-1806-06176}
Y.{-}H.~H. Tsai, P.~Liang, et~al.,
\newblock ``Learning factorized multimodal representations,''
\newblock in {\em ICLR}, 2019.

\bibitem{delbrouck-etal-2020-transformer}
J.~Delbrouck et~al.,
\newblock ``A transformer-based joint-encoding for emotion recognition and
  sentiment analysis,''
\newblock in {\em 2nd Challenge-HML}. 2020, pp. 1--7, ACL.

\bibitem{Sun2020LearningRB}
Z.~Sun et~al.,
\newblock ``Learning relationships between text, audio, and video via deep
  canonical correlation for multimodal language analysis,''
\newblock in {\em AAAI}, 2020.

\bibitem{yu2021le}
W.~Yu et~al.,
\newblock ``Learning modality-specific representations with self-supervised
  multi-task learning for multimodal sentiment analysis,''
\newblock in {\em Proc.AAAI}. 2021, arXiv.

\bibitem{kim_cmsbert-clr_2022}
J.~Kim and J.~Kim,
\newblock ``{CMSBERT}-{CLR}: {Context}-driven {Modality} {Shifting} {BERT} with
  {Contrastive} {Learning} for linguistic, visual, acoustic
  {Representations},'' 2022,
\newblock arXiv:2209.07424.

\bibitem{pfeiffer-etal-2021-adapterfusion}
J.~Pfeiffer, A.~Kamath, et~al.,
\newblock ``{A}dapter{F}usion: Non-destructive task composition for transfer
  learning,''
\newblock in {\em Proc. 16th ACL}. 2021, pp. 487--503, ACL.

\bibitem{9746418}
G.~Paraskevopoulos, E.~Georgiou, and A.~Potamianos,
\newblock ``Mmlatch: Bottom-up top-down fusion for multimodal sentiment
  analysis,''
\newblock in {\em ICASSP IEEE}, 2022, pp. 4573--4577.

\bibitem{tsai-etal-2019-multimodal}
Y.H.~H. Tsai, S.~Bai, P.~P. Liang, J.~Z. Kolter, L.P. Morency, and
  R.~Salakhutdinov,
\newblock ``Multimodal transformer for unaligned multimodal language
  sequences,''
\newblock in {\em Proc. 57th ACL}, 2019, pp. 6558--6569.

\bibitem{liu-etal-2018-efficient-low}
Z.~Liu et~al.,
\newblock ``Efficient low-rank multimodal fusion with modality-specific
  factors,''
\newblock in {\em Proc. 56th ACL}. 2018, pp. 2247--2256, ACL.

\bibitem{tensoremnlp17}
A.~Zadeh, M.~Chen, et~al.,
\newblock ``Tensor fusion network for multimodal sentiment analysis,''
\newblock in {\em EMNLP}, 2017.

\bibitem{Zadeh2018MultimodalLA}
A.~Zadeh, P.~P. Liang, et~al.,
\newblock ``Multimodal language analysis in the wild: Cmu-mosei dataset and
  interpretable dynamic fusion graph,''
\newblock in {\em ACL}, 2018.

\bibitem{DBLP:journals/corr/KingmaB14}
D.~P. Kingma and J.~Ba,
\newblock ``Adam: A method for stochastic optimization,''
\newblock in {\em ICLR}, 2015.

\bibitem{pennington2014glove}
J.~Pennington, R.~Socher, and C.~D. Manning,
\newblock ``Glove: Global vectors for word representation,''
\newblock in {\em EMNLP}, 2014, pp. 1532--1543.

\bibitem{hazarika2022analyzing}
D.~Hazarika, Y.~Li, et~al.,
\newblock ``Analyzing modality robustness in multimodal sentiment analysis,''
\newblock in {\em NAACL}. 2022, pp. 685--696, ACL.

\end{thebibliography}

\end{document}